\documentclass[conference,a4paper]{IEEEtran}
\IEEEoverridecommandlockouts

\usepackage[hidelinks]{hyperref}
\usepackage[cmex10]{amsmath}
\usepackage{amssymb,amsfonts}
\usepackage{dblfloatfix}

\usepackage[ruled,vlined]{algorithm2e}
\usepackage{graphicx}
\graphicspath{{figs}}

\usepackage{booktabs}
\usepackage{siunitx}
\usepackage[numbers,compress]{natbib}
\usepackage{texnames}
\usepackage{bm,bbm}
\usepackage{orcidlink}
\usepackage{subcaption}

\begin{document}

\title{\uppercase{D\textsuperscript{3}R-DETR: DETR with Dual-Domain Density Refinement for Tiny Object Detection in Aerial Images}
\thanks{\textit{Corresponding author: Yuhan Liu.}}
}

\author{	\IEEEauthorblockN{Zixiao Wen$^{1,2,3,4}$\orcidlink{0009-0008-3729-0538}, Zhen Yang$^{1,2,3}$, Xianjie Bao$^{1,2,3}$, Lei Zhang$^{1,2,3}$, \\
Xiantai Xiang$^{1,2,3,4}$, Wenshuai Li$^{1,2,3,4}$, Yuhan Liu$^{1,2,3,4}$}
	\IEEEauthorblockA{
		\textit{$^1$ Aerospace Information Research Institute, Chinese Academy of Sciences}\\
		100094 Beijing, China\\
	    \textit{$^2$ Key Laboratory of Technology in Geo-Spatial Information Processing and Application System, Chinese Academy of Sciences}\\
		100190 Beijing, China\\
		\textit{$^3$ Key Laboratory of Target Cognition and Application Technology, Chinese Academy of Sciences}\\
		100190 Beijing, China\\
		\textit{$^4$ School of Electronic, Electrical and Communication Engineering, University of Chinese Academy of Sciences}\\
		100049 Beijing, China\\
	    wenzixiao22@mails.ucas.ac.cn, yangzhen003999@aircas.ac.cn, baoxj@aircas.ac.cn, zhanglei@aircas.ac.cn, \\
		xiangxiantai@gmail.com, liwenshuai24@mails.ucas.ac.cn, liuyuhan@aircas.ac.cn}
}

\maketitle
\begin{abstract}
	Detecting tiny objects plays a vital role in remote sensing intelligent interpretation, as these objects often carry critical information for downstream applications.
	However, due to the extremely limited pixel information and significant variations in object density, mainstream Transformer-based detectors often suffer from slow convergence and inaccurate query-object matching.
	To address these challenges, we propose D\textsuperscript{3}R-DETR, a novel DETR-based detector with Dual-Domain Density Refinement.
	By fusing spatial and frequency domain information, our method refines low-level feature maps and utilizes their rich details to predict more accurate object density map, thereby guiding the model to precisely localize tiny objects.
	Extensive experiments on the AI-TOD-v2 dataset demonstrate that D\textsuperscript{3}R-DETR outperforms existing state-of-the-art detectors for tiny object detection.
\end{abstract}


\section{Introduction}\label{sec:intro}
Tiny object detection (TOD), which aims to locate and classify objects occupying extremely limited pixels (smaller than 16$\times$16 pixels~\cite{wang2021tiny}), is a critical task in remote sensing applications, including surveillance, environmental monitoring, and urban planning.
However, conventional feature enhancement methods struggle to address the challenges of missing or blurred object pixels, resulting in weak feature representations and making precise localization of tiny objects highly challenging. 
Moreover, the scenarios in remote sensing TOD datasets are highly diverse, covering a wide range of object types, from ships in open seas to vehicles in urban environments.
This leads to significant variations in object density, which further increases the risk of missed and false detections. 

To address the challenge of weak feature representation for tiny objects, researchers have explored the integration of frequency domain information to enhance feature expression.
HS-FPN~\cite{shi2025hs} combines high-frequency responses of object features with spatial features to strengthen feature maps at multiple scales.
SpectFormer~\cite{patro2025spectformer} replaces the standard multi-head self-attention module in Transformers with a frequency domain enhancement module.
FDA-IRSTD~\cite{zhu2025towards} improves the representation of infrared small targets by applying attention weighting to different frequency components in the feature spectrum.
FANet~\cite{wen2025fanet} further introduces frequency domain enhancement modules at both the feature map and RoI levels.
These approaches demonstrate the potential of frequency domain information in boosting the discriminative power of features for tiny object detection.
In addition, recent studies have introduced density map to guide the training of DETR-based detectors, aiming to improve query-object matching accuracy and object recall.
For example, DQ-DETR~\cite{huang2024dq} and D3Q~\cite{ye2025density} reconstruct density map from encoder memory and use them to dynamically generate queries with adaptive quantity and positions.
Dome-DETR~\cite{hu2025dome} designs a lightweight density-focal extractor to optimize both feature encoding and query selection.
DART~\cite{siddique2025dynamic} employs a density adaptive region attention mechanism to emphasize feature responses in high-density areas.

\begin{figure*}[!t]
    \centering
	\includegraphics[width=0.95\linewidth]{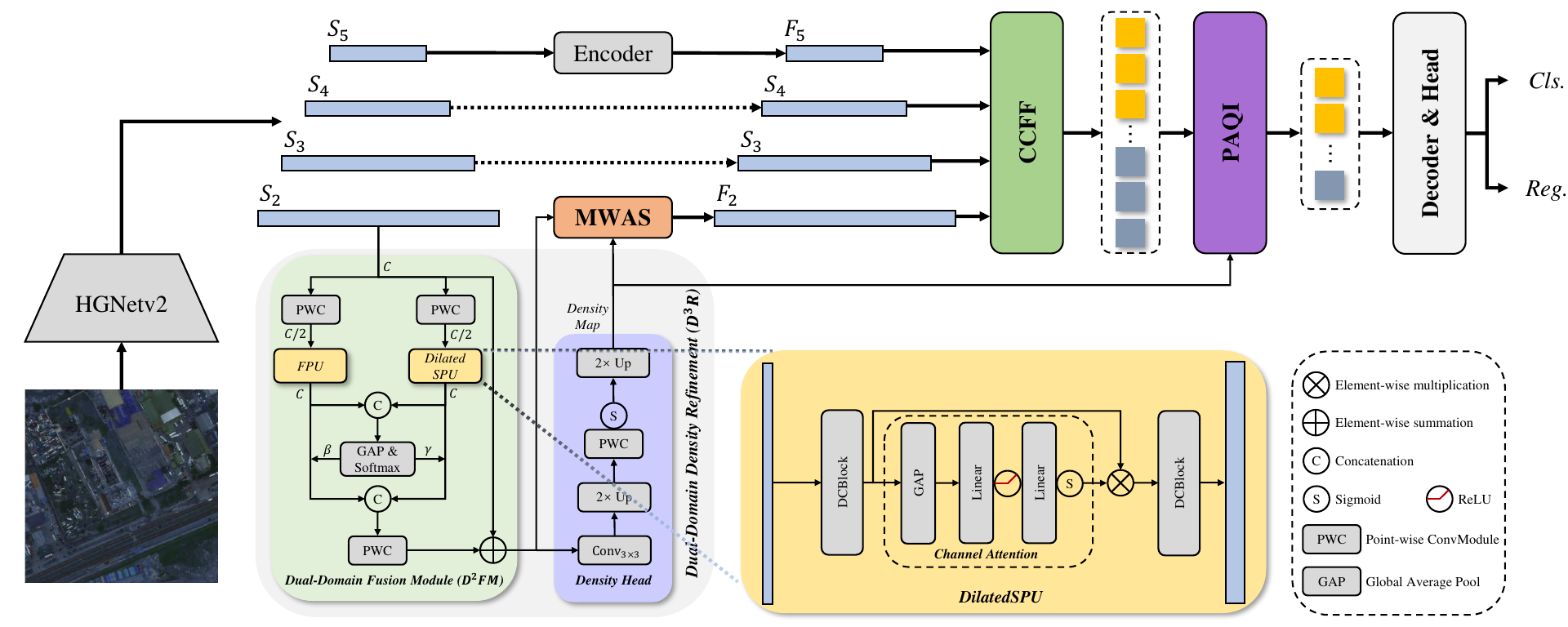}
    \caption{Overview architecture of our proposed D\textsuperscript{3}R-DETR.
	D\textsuperscript{2}FM fuses spatial and frequency domain information to extract richer features for accurate density map reconstruction, along with a lightweight density head.
	MWAS denotes Masked Window Attention Sparsification, and PAQI denotes Progressive Adaptive Query Initialization---both adopted from Dome-DETR~\cite{hu2025dome}.
	CCFF denotes CNN-based Cross-scale Feature Fusion~\cite{zhao2024detrs}.}
    \label{fig:overview}
\end{figure*}

Building on these advances, we propose a novel approach, named D\textsuperscript{3}R-DETR, which integrates Dual-Domain Density Refinement (D\textsuperscript{3}R) into the DETR framework.
Our method extends traditional density-guided frameworks by incorporating a Dual-Domain Fusion Module (D\textsuperscript{2}FM), which combines dilated convolution for spatial context modeling with filter kernels in the frequency domain.
This innovative design enables the extraction of richer and more detailed features, facilitating the reconstruction of more accurate object distribution representations.
Additionally, a lightweight density head is employed to guide the model to focus on high-density regions and support the generation of more precise queries for tiny object detection.
We conduct extensive experiments on the AI-TOD-v2 dataset to validate the effectiveness of our method.
The main contributions are as follows:
\begin{itemize}
	\item We propose D\textsuperscript{3}R-DETR, a novel DETR-based detector that incorporates D\textsuperscript{3}R method, guiding the model to focus on high-density regions.
	\item 	D\textsuperscript{2}FM is designed to fuse spatial and frequency domain information, along with a lightweight density head to reconstruct accurate density map to enhance feature representation and improve query-object matching for tiny object detection.
\end{itemize}

\section{Methodology}\label{sec:method}

\subsection{Overview}
As shown in Fig.~\ref{fig:overview}, our study introduces D\textsuperscript{3}R-DETR, which builds upon the Dome-DETR framework~\cite{hu2025dome}.
In this work, we incorporate the D\textsuperscript{3}R method, replacing the original Density-Focal Extractor (DeFE) with our proposed D\textsuperscript{2}FM and a lightweight density head.

\subsection{Dual-Domain Density Refinement}

\subsubsection{Dual-Domain Fusion Module}
\begin{figure}[hbt]
    \centering
	\includegraphics[width=0.95\linewidth]{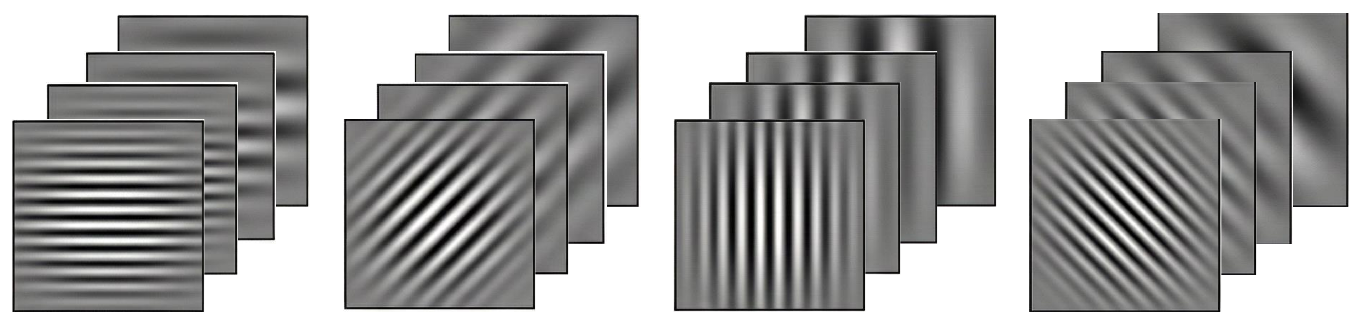}
    \caption{Visualization of FrGK in different angles and scales.}
    \label{fig:FrGK}
\end{figure}

The density map extractor in DeFE adopts a relatively simple approach, using only several layers of dilated convolution.
Although this increases the receptive field, it overlooks many fine details.
At the same time, the quality of the generated density map plays a crucial role in subsequent feature encoding and decoding.
Therefore, a more refined and detailed representation is necessary.
Inspired by SFS-Conv~\cite{li2024unleashing}, we design D\textsuperscript{2}FM, as shown in Fig.~\ref{fig:overview}.
The model utilizes FPU and DilatedSPU to extract spatial and frequency domain information, respectively.
The FPU applies Fractional Gabor Kernels (FrGK) for convolution, following~\cite{li2024unleashing}, formulated as:
\begin{align}
    F_{in} &= [F_{in}^{1}, F_{in}^{2}, \ldots, F_{in}^{N}] \\
    F_{mid}^n &= \mathrm{ConvBlock}(F_{in}^{n}, \mathrm{FrGK}),\ n=1,2,\ldots, N \\
    F_{out} &= \mathrm{PWC}\big(\mathrm{Concat}([F_{mid}^{1}, F_{mid}^{2}, \ldots, F_{mid}^{N}])\big)
\end{align}
where $N=4$, and $\mathrm{FrGK}$ contains Fractional Gabor Kernels with different angles and scales, as illustrated in Fig.~\ref{fig:FrGK}.
Here, $\mathrm{ConvBlock}(\cdot)$ denotes a composite operation consisting of convolution, activation, and pooling, and $\mathrm{PWC}(\cdot)$ denotes point-wise convolution with batch normalization and activation.
On the other hand, the DilatedSPU incorporates Dilated Convolution Block (DCBlock) and channel attention to enhance spatial feature modeling, as formulated below:
\begin{align}
    F_{mid}     &= \mathrm{DCBlock}_{1}(F_{in}) \\
    \hat{F}_{mid} &= \mathrm{CA}(F_{mid}) \odot F_{mid} \\
    F_{out}     &= \mathrm{DCBlock}_{2}(\hat{F}_{mid})
\end{align}
where $F_{mid}$ has $C/2$ channels and $F_{out}$ has $C$ channels. $\mathrm{CA}(\cdot)$ denotes the channel attention module, and $\odot$ represents the Hadamard product.

\begin{table*}[!t]
	\centering
	\caption{Comparison of the proposed D\textsuperscript{3}R-DETR with state-of-the-art method.
	* denotes a re-implementation of the results.}\label{tab:comparison}
	\begin{tabular}{lccccccccc}
		\toprule
		\textbf{Method} & \textbf{Source} & \textbf{Backbone} & \textbf{AP} & $\mathbf{AP_{50}}$ & $\mathbf{AP_{75}}$ & $\mathbf{AP}_{\bm{vt}}$ & $\mathbf{AP}_{\bm{t}}$ & $\mathbf{AP}_{\bm{s}}$ & $\mathbf{AP}_{\bm{m}}$ \\ \cmidrule(lr){1-1} \cmidrule(lr){2-2} \cmidrule(lr){3-3} \cmidrule(lr){4-6}\cmidrule(lr){7-10}
		ORFENet~\cite{liu2024tiny}	    & TGRS2024  & ResNet50   & 24.8 & 55.4 & 18.2 & 9.7  & 24.4 & 28.7 & 35.1 \\
		NWD-RKA~\cite{xu2022detecting}  & ISPRS2022 & ResNet50   & 24.7 & 57.4 & 17.1 & 9.7  & 24.2 & 29.8 & 39.3 \\
		RFLA~\cite{xu2022rfla}			& ECCV2022  & ResNet50 	 & 25.7 & 58.9 & 18.8 & 9.2  & 25.5 & 30.2 & 40.2 \\
		DINO-DETR~\cite{zhang2023dino}	& ICLR2023  & ResNet50 	 & 25.9 & 61.3 & 17.5 & 12.7 & 25.3 & 32.0 & 39.7 \\
		DQ-DETR~\cite{huang2024dq} 		& ECCV2024  & ResNet50 	 & 30.2 & \textbf{68.6} & 22.3 & 15.3 & 30.5 & 36.5 & 44.6 \\
		Dome-DETR*~\cite{hu2025dome} 	& ACMMM2025 & HGNetv2-B0 & 28.7 & 62.0 & 22.8 & 14.6 & 28.1 & 34.2 & 42.2 \\
		D\textsuperscript{3}R-DETR 		& Ours		& HGNetv2-B0 & \textbf{31.3 (+2.6)} & 65.1 & \textbf{26.2} & \textbf{16.6} & \textbf{30.8} & \textbf{36.8} & \textbf{44.7} \\ \bottomrule
	\end{tabular}
\end{table*}

\begin{figure}[hbt]
    \centering
	\includegraphics[width=0.75\linewidth]{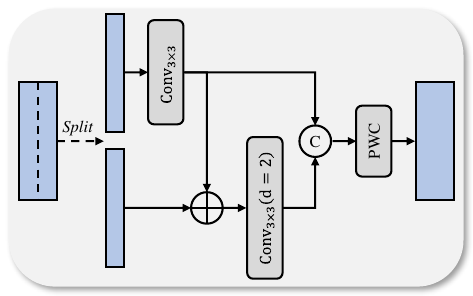}
    \caption{The proposed DCBlock in DilatedSPU.}
    \label{fig:DCBlock}
\end{figure}

To further illustrate the design and advantages of DCBlock, Fig.~\ref{fig:DCBlock} presents its detailed structure.
By leveraging dilated convolution and residual connections, DCBlock maintains high resolution and effectively integrates spatial information from different receptive fields. Specifically, DCBlock first splits the input feature channels into two groups, which are then processed by two 3$\times$3 convolutions with dilation rates (1,2).
Residual connections are employed to further expand the receptive field, allowing the extraction of multi-scale contextual information across different feature channels.
Finally, pointwise convolution is applied to achieve channel fusion.
This design significantly enhances the spatial feature representation capability of object distribution characteristics across various regions while introducing minimal computational overhead. 

\subsubsection{Lightweight Density Head}
To obtain a more accurate representation of object distribution, we design a lightweight density head composed of several convolution and upsampling layers.
This module transforms the output from D\textsuperscript{2}FM into a single-channel map, which is then used to guide the encoding in MWAS and the query generation in PAQI with the same configurations in~\cite{hu2025dome}. Meanwhile, we employ the Density Recall Focal Loss (DRFL)~\cite{hu2025dome} to constrain the reconstruction quality, ensuring that the result accurately reflects the distribution of objects.

\section{Results}

\subsection{Dataset and Implementation Details}
AI-TOD-v2~\cite{xu2022detecting} is a dataset for tiny object detection in aerial images, covering eight categories of common-seen tiny objects.
It contains 11214 training images, 2804 validation images, and 14018 test images, with 752745 annotated object instances.
The absolute object size of AI-TOD-v2 is only 12.7 pixels, with a standard deviation of 5.6 pixels, which poses significant challenges for tiny object detection.

All experiments are conducted on 4$\times$ NVIDIA RTX 4090 GPUs with a batch size of 4, using PyTorch 2.4.0 and CUDA 12.1.
To ensure stable convergence, we train the model for 120 epochs, followed by 25 epochs with and without advanced augmentation.  
During evaluation, we adopt the AI-TOD~\cite{wang2021tiny} benchmark metrics, including $\mathrm{AP}_{50}$, $\mathrm{AP}_{75}$, $\mathrm{AP}_{vt}$, $\mathrm{AP}_{t}$, $\mathrm{AP}_{s}$, and $\mathrm{AP}_{m}$.  
Other experimental settings are consistent with Dome-DETR-S~\cite{hu2025dome}, employing a 1-layer transformer encoder, a deformable transformer decoder, and HGNetv2-B0 as the CNN backbone for fair comparison.

\subsection{Comparison with state-of-the-art}

\begin{figure}[!b]
    \centering
    \begin{subfigure}{1.0\linewidth}
        \centering
        \includegraphics[width=0.75\linewidth]{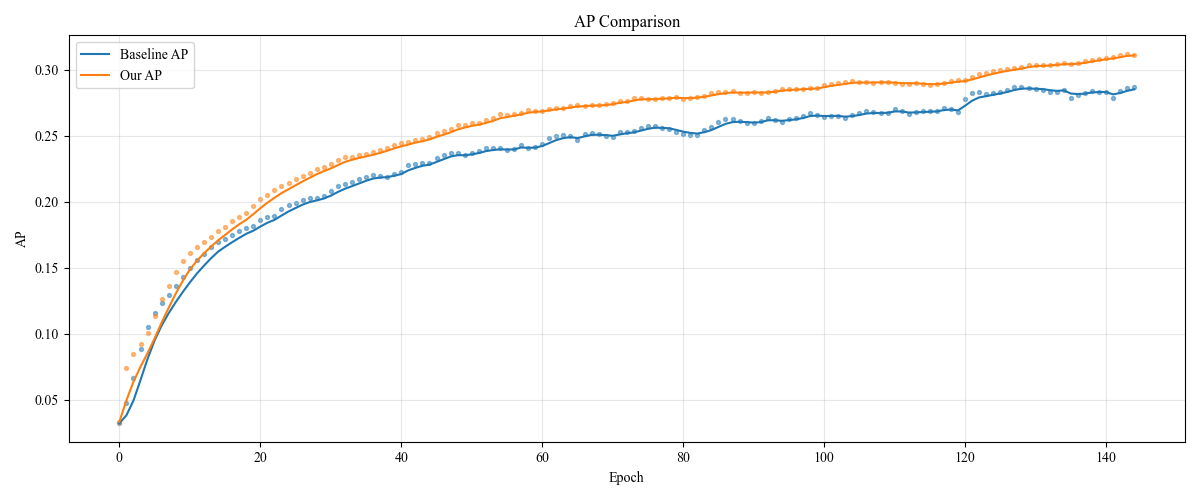}
        \caption{Average Precision (AP) performance comparisons.}
        \label{fig:ap}
    \end{subfigure}
    \vskip 0.5em
    \begin{subfigure}{1.0\linewidth}
        \centering
        \includegraphics[width=0.75\linewidth]{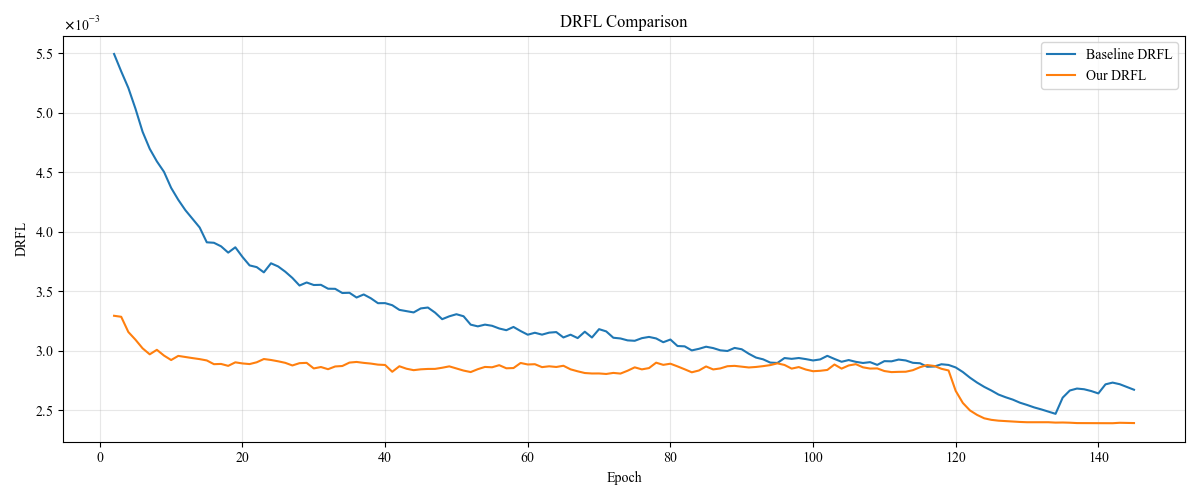}
        \caption{Density Recall Focal Loss (DRFL) comparisons.}
        \label{fig:drfl}
    \end{subfigure}
	\captionsetup{skip=1em}
    \caption{AP Performance and DRFL Comparisons.}
    \label{fig:comparison}
\end{figure}

\begin{figure*}[hbt]
    \centering
	\includegraphics[width=0.95\linewidth]{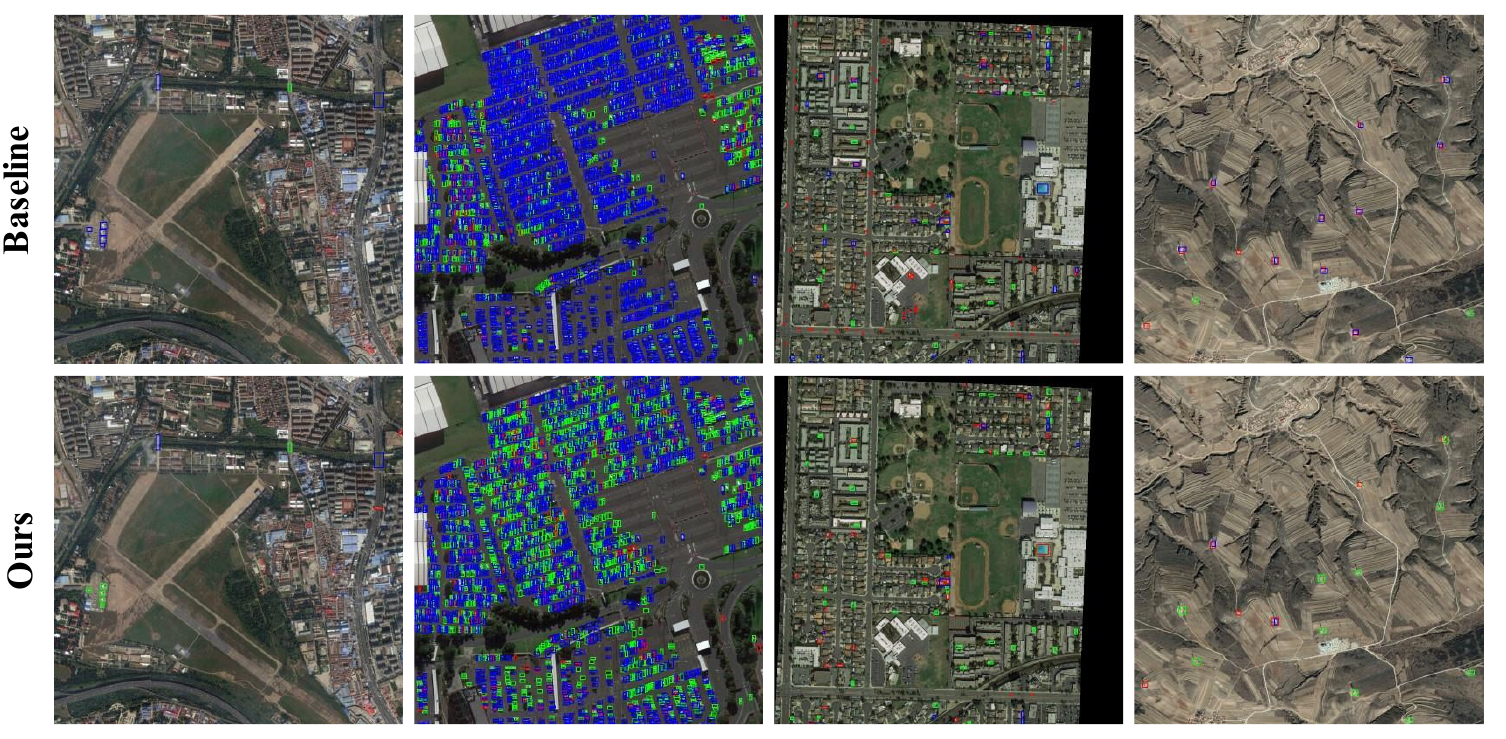}
    \caption{Qualitative results in AI-TOD-v2 test dataset.
    \textbf{Top row}: results of the baseline model;
    \textbf{Bottom row}: results of D\textsuperscript{3}R-DETR.
    The green, red, and blue boxes represent TP, FP, and FN, respectively.}
    \label{fig:results}
\end{figure*}

As shown in Table~\ref{tab:comparison}, we compare our proposed D\textsuperscript{3}R-DETR with existing state-of-the-art methods on the AI-TOD-v2 dataset, including CNN-based and DETR-based detectors.
The results demonstrate that D\textsuperscript{3}R-DETR outperforms all existing state-of-the-art methods on the AI-TOD-v2 dataset, and achieves significant improvements over the baseline model~\cite{hu2025dome}, with +2.6\% AP, +3.1\% $\mathrm{AP}_{50}$, +2.0\% $\mathrm{AP}_{vt}$ and +2.7\% $\mathrm{AP}_{t}$.
In addition, we compare the AP performance and DRFL loss convergence speed between D\textsuperscript{3}R-DETR and the baseline model to further validate the effectiveness of our feature extraction strategy.
As shown in Fig.~\ref{fig:comparison}, our model achieves notable performance improvements at different training stages and DRFL exhibits faster and more stable convergence.
These results indicate that leveraging dual-domain information enables more accurate modeling of object distributions, effectively guiding the model to focus on high-density regions.

Finally, we present qualitative results in Fig.~\ref{fig:results} to demonstrate the visual detection performance.
As shown in the figure, D\textsuperscript{3}R-DETR exhibits superior performance in detecting tiny objects in high-density regions, significantly reducing both missed detections and false positives.
These visual comparisons further validate that accurate density map reconstruction enables the model to better localize tiny objects, thereby enhancing overall detection performance.

\subsection{Ablation Study}
To further explore the effectiveness of frequency-domain information in D\textsuperscript{3}R-DETR, we conduct an ablation study to evaluate the effectiveness of different fractional filter kernels (FrFK) in the frequency domain processing of FPU: Garbor, Fourier, and Haar.
As shown in Table~\ref{tab:ablation}, the Garbor Kernels achieves the best performance with 31.3\% AP, demonstrating its superior capability in capturing frequency domain information for tiny object detection.

\begin{table}[hbt]
	\centering
	\caption{Detection performance of different FrFK.}\label{tab:ablation}
	\begin{tabular}{cccc}
		\toprule
		\textbf{FrFK} & \textbf{AP} & $\mathbf{AP}_{50}$ & $\mathbf{AP}_{75}$\\ \cmidrule(lr){1-1} \cmidrule(lr){2-4}
		baseline & 28.7 & 62.0 & 22.8 \\
        Haar     & 30.0 & 63.4 & 24.2 \\
        Fourier  & 30.3 & 63.8 & 24.7 \\
        Garbor	 & \textbf{31.3} & \textbf{65.1} & \textbf{26.2} \\
		\bottomrule
	\end{tabular}
\end{table}

\section{Discussion}
In this paper, we proposed D\textsuperscript{3}R-DETR, a novel detector designed for tiny object detection in aerial images.
By integrating the D\textsuperscript{3}R strategy, our method effectively addresses the challenges of weak feature representation and significant density variations inherent in tiny objects.
Specifically, the proposed D\textsuperscript{2}FM combines spatial context modeling via dilated convolution with frequency domain feature extraction using Convolutional Fractional Gabor Kernels.
This dual-domain approach enables the reconstruction of high-quality density maps, which in turn guide the model to focus on high-density regions and generate more precise queries.
Extensive experiments on the AI-TOD-v2 dataset demonstrate that D\textsuperscript{3}R-DETR achieves state-of-the-art performance, significantly outperforming existing methods.
In future work, we plan to further optimize detection performance by incorporating temporal and semantic information\cite{xiang2026slgnet}, enabling the model to better exploit contextual cues and improve robustness in more complex scenarios.

\small
\bibliographystyle{IEEEtranN}
\bibliography{references}

\end{document}